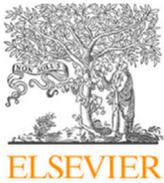
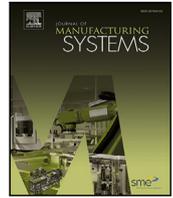

Technical paper

# Introducing PetriRL: An innovative framework for JSSP resolution integrating Petri nets and event-based reinforcement learning

Sofiene Lassoued *, Andreas Schwung

*Department of Automation Technology, South Westphalia University of Applied Sciences, Soest, 59494, North Rhine-Westphalia, Germany*

ABSTRACT

Resource utilization and production process optimization are crucial for companies in today's competitive industrial landscape. Addressing the complexities of job shop scheduling problems (JSSP) is essential to improving productivity, reducing costs, and ensuring timely delivery. We propose PetriRL, a novel framework integrating Petri nets and deep reinforcement learning (DRL) for JSSP optimization. PetriRL capitalizes on the inherent strengths of Petri nets in modelling discrete event systems while leveraging the advantages of a graph structure. The Petri net governs automated components of the process, ensuring adherence to JSSP constraints. This allows for synergistic collaboration with optimization algorithms such as DRL, particularly in critical decision-making. Unlike traditional methods, PetriRL eliminates the need to preprocess JSSP instances into disjunctive graphs and enhances the explainability of process status through its graphical structure based on places and transitions. Additionally, the inherent graph structure of Petri nets enables the dynamic additions of job operations during the inference phase without requiring agent retraining, thus enhancing flexibility. Experimental results demonstrate PetriRL's robust generalization across various instance sizes and its competitive performance on public test benchmarks and randomly generated instances. Results are compared to a wide range of optimization solutions such as heuristics, metaheuristics, and learning-based algorithms. Finally, the added values of the framework's key elements, such as event-based control and action masking, are studied in the ablation study.

## 1. Introduction

In today's highly competitive industrial landscape, it is crucial to efficiently utilize resources and optimize production processes to maintain a competitive edge. One of the most significant challenges in operations research and production planning is the Job Shop Scheduling Problem (JSSP). The JSSP involves the meticulous scheduling of jobs across a network of machines. Each job consists of a sequence of operations, which requires processing on a specific machine [1]. As industries aim to improve productivity, reduce costs, and ensure timely delivery, addressing the complexities of JSSP becomes a critical endeavour in achieving operational excellence.

In this pursuit, Petri nets emerge as a versatile tool for modelling and analysing systems characterized by concurrent processes, finding applications across diverse domains, including manufacturing, communication protocols, and software systems [2,3]. Particularly in job shop scheduling, Petri nets offer unique advantages. Their graphical representation helps to visualize and understand the scheduling process, making it more transparent. Moreover, they can function as a modelling and control tool, allowing them to be integrated with optimization algorithms. This integration leads to practical solutions for JSSPs [4].

Therefore, Petri nets are a promising approach to addressing the complexities of JSSP and enhancing operational efficiency in production planning.

Several techniques have been suggested to address JSSPs. Exact methods, as described in [5,6], provide optimal solutions but have drawbacks such as high computational demands and limited action space. On the other hand, approximate solutions, like the heuristics presented in [7,8], overcome computational challenges by simplifying complex tasks to converge to a solution quickly. Neural networks are iterative approaches offering approximate solutions. Compared to heuristics, neural networks rely less on domain knowledge, and their ability to learn intricate patterns and process large datasets has led to successful scheduling applications, as demonstrated in studies such as [9–11], and [12]. However, the "black box" nature of neural networks poses a significant challenge, hindering their widespread adoption in industries.

The contribution of this study can be summarized as follows:

1. We present PetriRL, a novel framework that utilizes Petri nets and actor–critic-based reinforcement learning to solve job-shop

   scheduling problems. This method enhances interpretability through Petri Nets' effective graphical representation.
2. We propose a flexible solution, thanks to the graph nature of the Petri net. This enables the real-time addition of operations into existing jobs during the inference process without retraining the agent.
3. We enhance efficiency by integrating event-based control and action space masking, limiting the needed agent–environment interaction to solve the JSSP.
4. We eliminate the requirement for laborious preprocessing and casting of the JSSP instances into disjunctive, reducing complexity and computation requirements and allowing efficient input of the raw instances in our system.
5. We conduct a comparative study on public benchmarks, comparing our results to heuristics, metaheuristics, and other learning-based approaches. Furthermore, we delve into ablation studies to analyse the contribution of individual elements of the framework.

The paper is structured as follows: Section 2 provides an overview of related works. This is followed by Section 3 in which preliminary concepts such as JSSP, Petri nets and Reinforcement learning are introduced. In Section 4, we establish the PetriRL framework, starting with the mathematical formulation of the JSSP using Petri nets and then the PetriRL environment. In Section 5, we introduce key elements of the proposed framework, including event-based control and action masking. Section 6 encompasses testing our approach on public benchmarks, results discussion, and an ablation study. Following this, we explore the generalization and flexibility of our solution. Section 7 serves as the conclusion, summarizing the paper and presenting future perspectives.

## 2. Literature review

One promising approach to tackling scheduling challenges involves employing Petri nets alongside optimization methods like heuristics, meta-heuristics, and iterative learning approaches such as reinforcement learning. In this literature review, we explore Petri nets' applications in production scheduling, followed by metaheuristics and iterative optimization tools such as reinforcement learning in solving JSSPs. Finally, we investigate the applications of graph neural networks before concluding with an identification of the research gap.

Petri nets are a powerful mathematical tool for modelling and analysing concurrent systems. Their benefits are widely recognized in the production industry, as they can efficiently represent multiple states, capture precedence relationships, model deadlocks, conflicts, and synchronization, among other critical aspects [2]. Extensions of Petri nets have further enhanced their capabilities. For example, Stochastic Petri nets allow for the representation and analysis of probabilistic elements in JSSP applications. Coloured Petri nets enable diverse types of tokens within a single net, resulting in a concise network [13,14]. Timed Petri nets incorporate time-related aspects, facilitating modelling time delays and temporal relationships, which are essential requirements for JSSP applications [15].

Petri nets are versatile tools used in various industrial applications and have three primary functions. Firstly, they can predict and manage system states, particularly in scenarios like deadlock prevention, through their reachability graphs [16]. Secondly, Petri nets have robust mathematical foundations that enable qualitative and quantitative systems analyses, providing valuable insights into their operational dynamics [17]. Lastly, Petri nets are indispensable simulation instruments that work together with optimization algorithms such as heuristics, metaheuristics, and iterative approaches to improve system efficiency and performance [18].

Heuristics are popular optimization tools in JSSP applications thanks to their simplicity and reduced computational requirements compared to exact solutions. However, these simplifications come at the cost of sub-optimal solutions and a heavy reliance on domain knowledge, hindering their generalization. Some meta-heuristics draw inspiration from nature, for example, Genetic Algorithms (GA) and Ant Colony Optimization (ACO), as discussed in [19,20], taking high-level strategies to solve JJSPs. While these approaches are not problem-specific and can be applied to a wide range of problems, they are still sensitive to initial conditions and require extensive tuning.

Reinforcement learning (RL) belongs to the iterative optimization tools. RL has proven effective in dynamic environments, As shown in the findings in [21], where 85% of evaluated RL implementations significantly improved scheduling performance in JSSPs. In addressing the agility requirements for the dynamic and flexible job shop scheduling problem (DFJSP), [22] implemented a deep Q-network (DQN) agent to select the best dispatching rule at each rescheduling point intelligently, another approach involved combining the duelling double Deep Q-network with prioritized replay (DDDQNPR) outperformed heuristic dispatching rules, particularly in dynamic environments with uncertain processing times [23]. Furthermore, [24] conducted experiments using multi-agent actor–critic techniques to solve JSSPs by applying the Deep Deterministic Policy Gradient (DDPG) algorithm. The authors of [25] have chosen an alternative method that utilizes hierarchical RL. This approach involves a top-level graph-based layer that breaks down the DFJSP into smaller, more manageable FJSP, which are then solved using an MLP. On the other hand, in [26], the FJSP are tackled using a multi-pointer graph network and multi-proximal policy optimization.

The graph-based approach gained prominence thanks to its ability to model complex relationships and adaptability to changing environments. [27] proposed a Graph Neural Network (GNN) embed the JSSP states; this resulted in a size-agnostic policy network effectively enabling generalization on large-scale instances. [28] leveraged GNN capability to comprehend the spatial structure of the problem, enabling adaptation to diverse JSSP instances and surpassing the performance of scenario-specific algorithms. Meanwhile, [29] introduced GraSP-RL to minimize makespan in a complex injection moulding production setting, achieving superior results to meta-heuristics like tabu search and genetic algorithms without additional training. A further optimization came from focusing on the most pertinent features instead of indiscriminate aggregation of information, [30] implemented a Gated-Attention model in which a gating mechanism is implemented to modulate the flow of learned features outperforming existing heuristics and state-of-the-art DRL baselines.

The success of the transformer-based models [31] in Large Language Models LLMs application led to their widespread across many applications. In paper [32], the authors used the node2vec algorithm to learn the JSSP instance's disjunctive graph characteristics in the form of a sequence, followed by a transformer architecture based on a multi-head attention mechanism to generate a solution. This approach offers long-range dependency thanks to the attention mechanism and parallel computing capacities vital to solving large-scale problems.

After reviewing the existing literature, it has become evident that an interpretable modelling tool tailored to JSSP environments that seamlessly integrates with a Deep Reinforcement Learning optimizer is needed. Incorporating Petri nets into reinforcement learning offers a promising avenue to address this gap. This approach eliminates the need to convert JSSP instances into disjunctive graphs, thus reducing complexity and computation. Moreover, by using an actor–critic agent alongside the Petri net, the resolution of JSSPs could be achieved with improved sample efficiency due to action masking and event-based control. In the following section, we present PetriRL, a framework for JSSP resolution that leverages Petri nets and actor–critic reinforcement learning.

## 3. Preliminaries

### 3.1. Job shop scheduling problem

JSSP is a classic optimization problem in operations research. It involves scheduling a set $\mathcal{J} = \{j_1, \ldots, j_J\}$ of $J$ jobs on a set $\mathcal{M} = \{m_1, \ldots, m_M\}$ of $M$ machines. Each job $j_i$ consists of a specific sequence of $ni$ consecutive operations $O_i = \{O_{i1}, O_{i2}, \ldots, O_{ini}\}$ with precedence constrain. Each operation $O_{ij}$ is associated with processing time $p_{ij}$ and a machine $m \in \mathcal{M}$. Once initiated, each operation remains uninterrupted, and a single machine can execute only one operation concurrently. Besides, all machines and jobs are available at the start of the time horizon. In this work, the optimization goal of JSSP is to determine a sequence of operations on each machine to minimize the makespan $C_{max}$ described by Eq. (1), where $C_i$ denotes the completion time of job $i$. Let $J \times M$ denote the size of a JSSP instance.

$$C_{max} = max\{C_i\}, i \in \{1, \ldots, J\} \quad (1)$$

### 3.2. Petri nets

Petri nets are mathematical models used to represent and analyse dynamic systems with concurrency and synchronization. They provide a graphical and mathematical framework for understanding system behaviour, enabling analysis of properties like reachability, liveness, and deadlock avoidance. Petri nets can be described as a pair $(\mathcal{G}, \mu_0)$, where $\mathcal{G}$ is a directed bipartite graph consisting of a finite set of nodes and edges, and $\mu_0$ represents the initial marking, which indicates the distribution of tokens among the places of the Petri net at its initial state. Compared to ordinary graphs, Petri nets have a distinct structure. Nodes in Petri nets are divided into two disjoint partitions: places $\mathcal{P}$ and transitions $\mathcal{T}$. Places represent system states or conditions, while transitions represent events or actions that change the system's state. The edges of the graph, also known as arcs, define the flow of tokens between places and transitions, indicating the conditions necessary for a transition to fire.

For any node n in $\mathcal{P} \cup \mathcal{T}$, we define $\pi(n)$: the set of upstream nodes, $\sigma(n)$: the set of downstream nodes, and $\mu(p)$ is the marking. When a transition $t \in \mathcal{T}$ fires the new marking is given by:

$$\tilde{\mu}(p) = \begin{cases} \mu(p) - 1 & \forall \ p \in \pi(t) \\ \mu(p) + 1 & \forall \ p \in \sigma(t) \\ \mu(p) & \text{otherwise} \end{cases} \quad (2)$$

Coloured Petri nets (CPNs) extend the basic Petri net model by allowing tokens to carry additional information, known as colours or attributes, for a higher level of expressiveness. CPNs offer the advantage of folding processes with similar structures but with different properties in a compact manner [33]. This proves especially advantageous in applications related to JSSPs, where different jobs share the same shop floor. A marked-coloured Petri net is defined by:

CPN = $(\mathcal{P}, \mathcal{T}, \mathcal{A}, \Sigma, C, N, E, G, I)$

with:

- $\mathcal{P} = \{p_1, p_2, \ldots, p_n\}$ : set of places,
- $\mathcal{T} = \{t_1, t_2, \ldots, t_n\}$ : set of transitions,
- $\mathcal{A} = \{a_1, a_2, \ldots, a_n\}$ : set of arcs,
- $\Sigma = \{c_1, c_2, \ldots, c_n\}$ : set of colours,
- $C : T \cup P \to \Phi(\Sigma), \Phi(\Sigma) \subset \Sigma$ is the colour function,
- $N : A \to (P \times T) \cup (T \times P)$ is the node function,
- $E : A \to e$ arc expression function,
- $G : T \to \{0, 1\}$ is the transitions guard function,
- $I : P \to$ initiation sequence is the initialization function.

#### 3.2.1. Petri nets versus disjunctive graphs

Although disjunctive graphs are frequently employed in the literature on solving JSSPs, this manuscript advocates for the utilization of Petri nets due to three primary reasons: their inherent explainability, their compatibility with DRL techniques, and their yet unexplored potential.

Petri nets offer a distinct advantage over disjunctive graphs by providing enhanced explainability. Their structure, organized around transitions and places, allows for clear differentiation between conditions and events. Places represent conditions such as the availability of resources or idle machines, while transitions signify events like the allocation of resources. The distribution of tokens within the Petri net provides an immediate snapshot of the system state. For instance, tokens in machine places indicate active processing, while tokens in job and delivery places represent remaining work and completed tasks, respectively. Based on the marking in the upstream places the Petri nets's guard function determines whether the transition is enabled. This capability to identify enabled transitions in real-time is particularly pertinent when employing Maskable Proximal Policy Optimization (MPPO), as discussed in Section 5.2.

Disjunctive graphs, a specialized graph type, provide a modelling approach for JSSPs. They are employed in resolving machine sequencing challenges within JSSPs by identifying the mini-max path within the graphs [34]. To feed these graphs to neural network optimization algorithms, additional methodologies such as node2vec [35] are essential for extracting meaningful feature vectors. In contrast, Petri nets's transitions and marking enable direct interactions with reinforcement learning agents. The token distributions constitute a valuable part of the observation, and each transition firing acts as an action within the agent's action space, capable of changing the system state.

Finally, the literature review has revealed an unexplored potential in integrating Petri nets with DRL to tackle JSSPs, making it an exciting research topic. In Addition, the extensive body of literature on Petri net analysis is a valuable resource that can be utilized, including readily available various mathematical analysis tools.

### 3.3. Reinforcement learning

Reinforcement learning (RL) is a field of machine learning that focuses on training agents to make sequential decisions in a given environment to maximize cumulative rewards. RL is based on the framework of Markov Decision Processes (MDPs), which consist of a tuple $(\mathcal{S}, \mathcal{A}, P, R, \gamma)$. Here, $\mathcal{S}$ represents the set of possible states, $\mathcal{A}$ is the set of available actions, $P$ denotes the state transition probability function $P(s'|s,a)$ indicating the likelihood of transitioning to state $s'$ from state $s$ by taking action $a$. $R$ signifies the reward function $R(s, a, s')$, which provides the immediate reward upon transitioning from state $s$ to state $s'$ through action $a$. The parameter $\gamma$ is the discount factor, determining the balance between immediate and future rewards in the expected return. Reinforcement learning algorithms aim to learn an optimal policy $\pi^*$ that maximizes the expected cumulative discounted rewards $G_t = \sum_{k=0}^{\infty} \gamma^k R_{t+k+1}$. By interacting with the environment, the algorithm tries to learn the optimal policy that results in the maximum possible reward over time [36].

## 4. PetriRL: a JSSP solving framework

This section starts by mathematically formulating the JSSPs using a coloured Petri net. Then, we construct an RL environment by introducing key elements of the RL framework, such as observation, action space, and reward shaping.

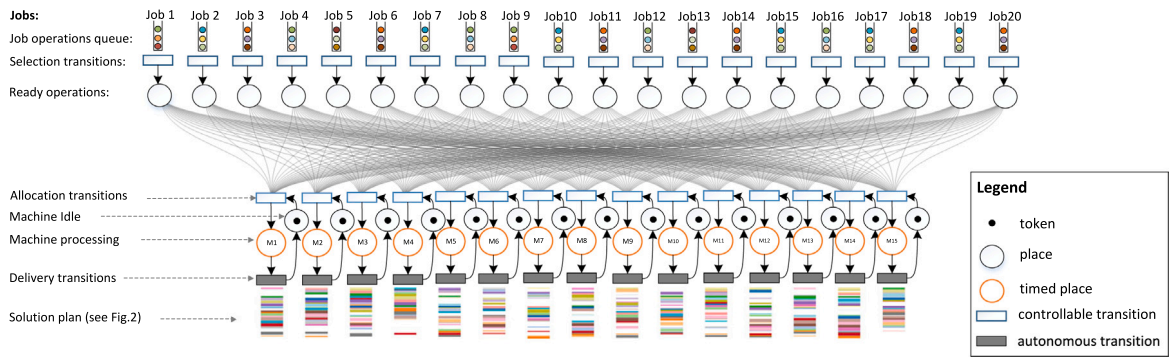

**Fig. 1.** PetriRL framework graphical representation example for a 20 jobs 15 machines shop floor.

*4.1. Petri net model of a JSSP*

After presenting the generic coloured Petri net model in the preliminary section, we customize the formulation to fit the JSSP study case. The set of all places is the union of job places $\mathcal{P}_j$, ready operations places $\mathcal{P}_r$, machine places $\mathcal{P}_m$, and idle places $\mathcal{P}_i$. The set of all transitions is the union of selection transitions $\mathcal{T}_s$, allocation transitions $\mathcal{T}_a$, and delivery transitions $\mathcal{T}_d$.

- $\mathcal{P} = \mathcal{P}_j \bigcup \mathcal{P}_r \bigcup \mathcal{P}_m \bigcup \mathcal{P}_i : j, r \in (1 \dots J) \quad m, i \in (1 \dots M)$
- $\mathcal{T} = \mathcal{T}_s \bigcup \mathcal{T}_a \bigcup \mathcal{T}_d : s, a \in (1 \dots J) \quad d \in (1 \dots M)$
- $\Sigma = \{c_1, c_2, \dots, c_M\} :$
- $C : \mathcal{T}_a \rightarrow \Phi(\Sigma) :$

Depending on whether the transition is autonomous, such as deliver transitions $\mathcal{T}_d$, controllable, such as selection transitions $\mathcal{T}_s$, or a combination of controllable and coloured transitions, such as the case of allocation transitions $\mathcal{T}_a$, the transition guard function **G** is given by:

$$G_d(t \in \mathcal{T}_d) = \begin{cases} 1 & , \text{if } \forall\ p \in \pi(t) \mid \mu(p) \geq 1 \\ 0 & , \text{otherwise} \end{cases} \quad (3)$$

$$G_s(t \in \mathcal{T}_s) = \begin{cases} 1 & , \text{if control signal = True} \\ & \wedge\ (\ \forall\ p \in \pi(t) \mid \mu(p) \geq 1\ ) \\ 0 & , \text{otherwise} \end{cases} \quad (4)$$

$$G_a(t \in \mathcal{T}_a) = \begin{cases} 1 & , \text{if control signal = True} \\ & \wedge\ (\ \exists\ p_i, p_r \in \pi(t) \mid \mu(p_i) \wedge \mu(p_r) \geq 1\ ) \\ & \wedge\ (\ token \in \bigcup_{p \in \pi(t)} \text{Tokens}(p),\ C(token) = C(t)\ ) \\ 0 & , \text{otherwise} \end{cases} \quad (5)$$

Finally, the initiation function **I** is extracted directly from the JSSP instance. Every job operations sequence in the instance is converted into an ordered list of tokens. The tokens bear the colour of the destination machine on top of additional features such as the job it belongs to and the processing time.

*4.2. The PetriRL environment*

After establishing the mathematical groundwork, the next step is constructing the RL environment. This involves initially providing a broad overview of the environment and a detailed breakdown of its elements, such as observations, action space, and credit attribution.

*4.2.1. Environment description*

In the proposed setting, as illustrated in Fig. 1, various concepts from Petri net formalism are applied to model a job shop environment to address JSSPs. This environment can be segmented into three main components: input, processing, and delivery.

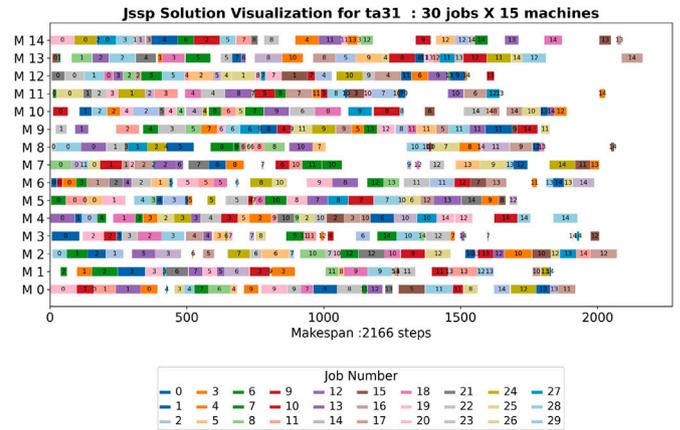

**Fig. 2.** Planning solution for Taillard benchmark instance "ta31": 30 jobs × 15 machines.

In the input phase, each job comprises an ordered sequence of operations converted into tokens. The operation tokens bear the colour of the destination machine and features such as processing time and the job they belong to. The job operations queue is connected to a controllable selection transition $t \in \mathcal{T}_s$. The RL agent triggers a selection transition to select a given job queue. Subsequently, the first operation token in the queue is transferred to the ready operations' place and is waiting to be allocated to a machine.

The processing section comprises the allocation transition, machine processing places, and corresponding idle check places. The allocation transition $t \in \mathcal{T}_a$ is colour-specific, allowing it only to consume tokens with a compatible colour. This ensures that the job operation is processed on the correct machine. Machine idle check-places act as safeguards, ensuring a machine can only process one operation simultaneously. The allocation transition is enabled and triggered by the RL agent only if it satisfies the guard function $G(t \in \mathcal{T}_a)$ aka a token with the correct colour is available in one of the ready operations place, and the machine is idle. The operation token is transferred to the machine processing location upon firing the allocation transition. Since the machine processing place is timed, the token must spend the processing time specified by its processing time before becoming available for the delivery transition.

Finally, the delivery process is automated, with autonomous delivery transitions $t \in \mathcal{T}_d$ firing promptly upon the availability of tokens. Once fired, the token is moved from the processing place to the delivery place. The environment has an internal clock that governs all time-related aspects such as processing duration, idle duration, job queue waiting duration, remaining times, machine entry date, token logging, etc. Logging is saved in the token history for every token transfer from one place to another. Ultimately, by extracting the entry and departure times of the token in the processing places, the solution plan can be derived, as illustrated in Fig. 2.

Following the above-mentioned concept, we have developed a custom Gym environment [37]. In the upcoming sections, we elaborate on various aspects of this environment, including observation, termination condition, action space, and rewards. The environment is publicly available as a package on the official repository for Python packages under the name **jsspetri**.

*4.2.2. Observation*

The observation provides the agent with information about the current environment state. A state represents the relevant aspects of the environment that the agent needs to consider when making decisions. The quality and completeness of observations directly impact the agent's ability to understand and learn from the environment.

The observation is represented as a one-dimensional vector with three components. First, it includes information about each machine's current state, represented by the remaining processing time. Second, the waiting queue for job operations provides information about the remaining work. The "observation depth" is a hyperparameter that determines the number of future operations to be included in the observation. If the depth is set to one, the agent will focus only on the job operations in the first row of each job queue. As the observation depth increases, the agent looks further into the future. Finally, the number of delivered pieces is also included in the observation. Termination is determined by confirming that all job queues and machines are empty.

*4.2.3. Action space*

The agent's margin of manoeuvre defines the action space. In the proposed framework, the agent role is limited to allocating a selected job operation to a specific processing machine. Once the job is allocated, the remaining is automated. The delivery transition operates autonomously, triggering promptly upon the token's availability following the token finishing the required processing duration in the machine. This simultaneously delivers the piece and flags that the machine is now idle. Since the allocation transition is coloured, the allocation transition is enabled if and only if a token with a matching colour is waiting in the ready operations and the machine is idle, as dictated by the transition guard function Eq. (5).

The choice of adding a standby action is a delicate one. Adding a standby option is necessary to allow the agent to opt for long-term strategies, such as leaving a machine idle for a given time to allocate a better option in the future. This also comes with the risk of the agent always choosing the standby action as a default choice in a reward hacking problem. This was solved in Section 5.2, where we discuss dynamic action space masking.

In summary, the action space encompasses all combinations of jobs and machine transitions, with the addition of the standby action. Its size is subject to dynamic adjustments based on the action mask, which selectively enables specific actions based on the Petri net's transition guard function.

*4.2.4. Reward shaping*

The reward function is one of the cornerstones in the reinforcement learning paradigm. The reward plays a crucial role in shaping the RL agent policy by offering a guiding signal, which the agent iteratively uses to improve the policy. Sparse reward is a situation where the agent does not receive frequent feedback from the environment, which poses a challenge for learning. In this case, the agent might struggle to map the actions with the outcome, hindering the capacity to form a good policy. To address the issue, researchers and practitioners often employ reward shaping, curriculum learning, or experience replay [38,39] to increase the feedback frequency synthetically.

In our JSSP case study, the selected performance metric and reward function are centred around minimizing the makespan, which is the time required to complete all predefined jobs. Since makespan inherently constitutes a sparse reward, as it only provides feedback at the end of task completion, challenges related to sparse rewards emerge. To address this issue, we opt for reward shaping. However, this solution brings drawbacks, including a loss of generalization, a reliance on domain expertise, and the potential for failure in dynamic environments. Additionally, if the reward function is too specific, there is a risk of reward hacking, where the agent may find shortcuts to maximize reward but deviate from the true objectives of the process.

Solving the problem involves maintaining a broad objective while increasing the feedback frequency. This method helps avoid overfitting, retain generalization properties, and ensure consistent feedback. We can rely on machine utilization as a reward metric to increase the feedback frequency. The reward can be calculated using the given formula:

$$\text{Reward} = 1 - \left( \frac{\sum \text{idle machines}}{\text{total number of machines}} \right) \tag{6}$$

We can apply a penalty when the standby action is chosen to discourage unnecessary idleness and penalize standby states while allowing for a strategic standby action choice. For instance, a penalty of $-0.1$ can be deducted from the reward when the standby action is selected.

## 5. Event-based control and action masking

In this section, we present two essential components of our PetriRL framework that significantly enhance the efficiency of our algorithm. The first component is event-based control, which ensures that token allocation is performed on demand, considerably improving sample efficiency. The second component is action masking, ensuring only valid actions are executed, thereby enhancing credit assignment. We illustrate how our framework's various elements interact in Fig. 3. The Petri net's dynamic guard function plays a crucial role in providing the dynamic mask for the Maskable PPO and acts as the trigger for event-based control. We conclude this section by enumerating some of the advantages of our framework compared to other approaches, notably removing the requirement for input preprocessing.

*5.1. Event-based control*

Event-based control refers to a paradigm where learning updates are triggered by specific events or occurrences rather than a fixed time interval [40,41]. In traditional RL, updates often occur at regular intervals or after completing a specific number of steps. In contrast, event-based control adapts its learning schedule based on events in the environment. This approach can be advantageous in dynamic and unpredictable environments where the occurrence of important events is not uniform over time. In our case study, the environment and agent interaction is only triggered if one machine or more is available.

Our approach involves on-demand agent intervention. When all machines are in use, or no operation can be assigned to an idle machine, the algorithm remains within the environment, incrementing only the internal clock. If at least one machine is available, it prompts the agent to allocate a job to the idle machine or choose to stand by. The standby option is only possible if an idle machine is present, allowing the agent to strategically opt for long-term planning without making the standby action the default choice. Moreover, decoupling the agent's step count from the environment's internal step clock offers two primary advantages: improved reward assignment and more efficient utilization of environment–agent interactions.

First, reward is one of the critical parts of reinforcement learning. Event-based control plays a pivotal role in achieving appropriate credit assignments. Without event-based control, the agent faces the challenge of consistently incurring negative rewards when all machines are occupied. This leads to agent confusion and dilutes the infrequent occurrence of positive rewards. The second advantage of

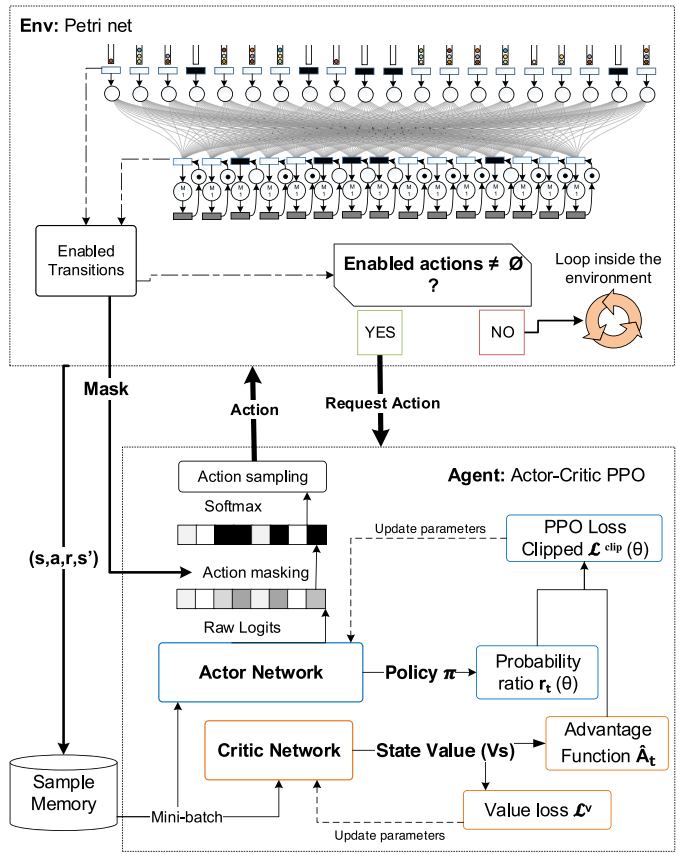

Fig. 3. Illustration of the interaction between the various elements of the PetriRL framework. The top box represents the environment depicted by the Petri net, while the lower box depicts the agent, a Maskable Proximal Policy Optimization algorithm.

**Algorithm 1** Proximal Policy Optimization with Action Masking using Petri nets' guard function.

1: Initialize policy parameters $\theta$, value function parameters $\phi$
2: Set hyperparameters, including the clipping parameter $\epsilon$
3: **for** iteration = 1, 2, … **do**
4:     Collect trajectories using the current policy $\pi_\theta$
5:     Compute advantages $\hat{A}_t$ and returns $R_t$
    **Update the Value Function:** State Compute the value function loss:

$$L^{VF}(\theta) = \mathbb{E}\left[\left(V_{\theta_{old}}(s_t) - V_\theta(s_t)\right)^2\right]$$

    State Optimize the value function parameters $\theta$ State Compute the Mask using the Petri net's guard function:

$$\text{Mask}[i] = G(a_i) \quad \forall a_i \in \mathcal{T}_a$$

    State Compute the actor's policy logits:

$$z_t(a_t) = \begin{cases} \text{policy\_network}(s_t, a_t) & \text{if } G(a_t) = 1 \\ -\infty & \text{else} \end{cases}$$

    State Convert the logits into a probability distribution over actions and calculate the action probability ratio:

$$\pi_\theta(a_t|s_t) = \sum_{a'} \frac{e^{z_t(a)}}{e^{z_t(a')}}, \quad \rho_t = \frac{\pi_\theta(a_t|s_t)}{\pi_{\theta_{old}}(a_t|s_t)}$$

6:     Compute Clipped surrogate objective:

$$L^{CLIP}(\theta) = \mathbb{E}_t\left[\min\left(\rho_t \hat{A}_t, \text{clip}(\rho_t, 1-\epsilon, 1+\epsilon)\hat{A}_t\right)\right]$$

7:     The final PPO loss ($C_1$ is a balancing hyper-parameter):

$$L(\theta) = L^{CLIP}(\theta) + C_1 L^{VF}(\theta)$$

8:     Optimize the policy parameters $\theta$ ($\alpha$ is the learning rate):

$$\theta \leftarrow \theta - \alpha \nabla_\theta L(\theta)$$

9: **end for**

employing event-based control is the optimal utilization of the available training environment–agent interactions. With event-based control implemented, the agent engages with the environment on demand, ensuring that the agent receives a relevant reward. This guarantees the efficient use of every training step, in contrast to the non-event-based approach, where a significant portion of interactions may be wasted on selecting the standby action due to all machines being occupied. Empirical results validated this observation. When comparing the total time steps during training with the internal environment clock, it is found that only 9% of the steps are utilized for allocation decision-making.

In summary, adopting event-based control significantly reduces the required agent–environment interactions to solve the JSSP. This is achieved by prompting the agent to make decisions on demand whenever a machine is idle. The benefits extend beyond enhanced learning due to improved credit assignments and are more efficient. When combined with the action masking strategy detailed in the subsequent paragraph, our approach ensures the effective utilization of available agent–environment interactions.

*5.2. Maskable proximal policy optimization*

Reinforcement learning excels in training agents to make sequential decisions, but the sheer number of possible actions in complex environments can hinder learning efficiency. Action masking is a vital technique to address this challenge [42]. The authors of the paper [43] outlined four strategies for managing invalid actions: a control setup where no action is used, assigning an action penalty, applying naive action masking, or using action masking. The First strategy involves assigning a penalty upon choosing an invalid action, such as implemented in [44]. In the second approach, called naive action masking, the action is removed directly from the action space without directly modifying the policy. Since the action is not chosen after its exclusion, the gradient gradually diminishes the future probability of the invalid action. The last strategy is action masking. In this approach, actions are masked by directly manipulating the raw outputs of the policy, also called logits, before Softmax normalization. By assigning a significantly negative value to the logits of the action to be masked, the probability of the invalid action becomes negligible.

The findings in [43] assert that masking removal is possible only in small action spaces, a finding also confirmed by our results in the ablation study. Naive action masking, however, exhibits drawbacks such as an elevated average Kullback–Leibler (KL) divergence between the target and current policy, leading to training instability and a volatile, sensitive behaviour. Lastly, the invalid actions penalty proves impractical for scalability; the results indicate a performance collapse with an increase in the environment size compared to action masking. Furthermore, setting the penalty value for an invalid action is a challenging hyper-parameter to tune and integrate with the overall environment reward. Due to these considerations, we choose to act on the policy as an invalid-actions handling policy.

The role of the Petri net in this framework goes beyond a simple environment that provides the RL agent with states and rewards. The Petri net also provides the mask used by the Maskable PPO agent to update the policy. The mask is calculated using the Petri net's guard function $G(t \in \mathcal{T}_a)$ (5) for the allocation transitions. The mask is a 1-D binary vector given by:

$$\text{Mask}[i] = G(a_i) \quad \forall a_i \in \mathcal{T}_a \text{ and } i \in (1 \ldots J) \tag{7}$$

The Petri net works in synergy with the RL agent to solve the JJSP. The corresponding pseudocode can be found in Algorithm 1.

Event-based control and action space masking share similarities. Both approaches aim to eliminate unnecessary steps and promote efficacy and performance. Like event-based control, eliminating unnecessary agent–environment interaction, action masking prevents training steps from being wasted on non-valid actions. We monitored the average frequency of valid actions within our environment to gauge the advantages of employing these techniques. We found that at any given moment, only a fraction of 10% of the actions are enabled. This phenomenon can be explained by the synchronization property of Petri nets. This means that for a transition firing to occur, multiple conditions must be met simultaneously. For instance, to permit the allocation transition firing, the machine must be idle, and there should be a compatible operation in the job operations waiting queue. This explains the low ratio of enabled action. If not implemented, the agent will waste nine out of ten interaction steps on non-valid action and receive a penalty. Incorporating action masking leads to a significant performance enhancement when paired with event-based control. This notable improvement is demonstrated in our ablation study.

### 5.3. Approach advantages

The suggested environment presents several advantages, notably better explainability and eliminating the need for extensive preprocessing. In many of the competing learning-based reinforcement learning approaches to solve the JSSP, preprocessing is required [27,30,32]. This preprocessing involves transforming each JSSP instance into a disjunctive graph representation. Once represented in the graph format, feature extraction techniques are utilized, including methods like "node2vec" [35] and attention mechanisms. These techniques aim to extract features that are more tailored to the characteristics of the employed optimizer. Unless the used optimizer is a graph neural network, this additional step could be omitted. The disjunctive graph representing the JSSP contains two types of information. First, structural and routing information is encoded into the edges of the disjunctive graphs. Second, the nodes contain features such as processing times. Since the structure is shared between all the jobs, the Petri net's graph nature can be leveraged to encode the structural and routing part of the input. This allows us to remove the need to transform JSSPs into disjunctive graphs and replace them with a list of coloured tokens. Compared to disjunctive graphs, representing JSSPs with a list of coloured tokens offers a more efficient, flexible, and streamlined approach.

Another advantage of Petri nets is their capability to serve as a comprehensive end-to-end control solution. Our approach consisted of delegating a fraction of the control to the Petri nets by utilizing tools such as timed places, autonomous transitions, and synchronization to streamline the automated parts of the process. This allows the agent to concentrate solely on essential decision-making and allocation tasks through the controlled transitions. In essence, the agent delegates automated sections of the process to the Petri net, enabling more focus on decision-making tasks. This balanced share of control is a key element for efficiency and explainability.

## 6. Results and discussion

### 6.1. Training

The algorithm underwent testing using public Taillard Benchmark instances [45]. These instances range in size from 15 × 15 to 100 × 20 (jobs × machines). The benchmark consists of eight groups, each comprising ten instances of identical sizes but with varying values, resulting in a total of 80 instances. We used the first instance in each group for training the agent and the remaining nine instances for testing. We utilize the MPPO (maskable policy proximal optimization) algorithm from the sb3-contrib library, a part of the openAi stable baselines3 [46]. The openAi stable baselines3 is a reliable implementation of reinforcement learning algorithms in PyTorch.

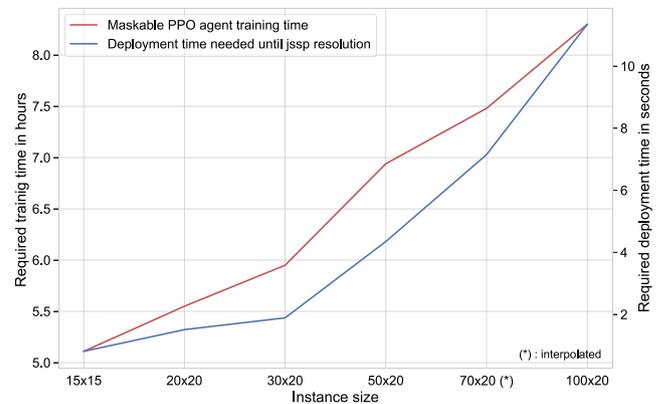

**Fig. 4.** Training and deployment time for different instances using PetriRL framework. The red line represents the agent training time requirement for different instance sizes, the corresponding *y*-axis on the left, and the blue line represents the time requirement to solve a JSSP during the inference phase, the corresponding *y*-axis to the right. (For interpretation of the references to colour in this figure legend, the reader is referred to the web version of this article.)

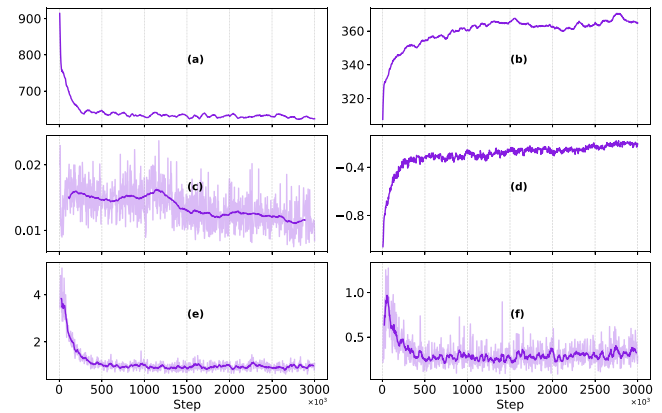

**Fig. 5.** Agent training performances, for instance "ta41"(30x 20). (**a**) the episode length,(**b**) the episode mean reward,(**c**) the Kullback–Leibler divergence evolution,(**d**) the entropy loss,(**e**) the value function loss,(**f**) the overall loss) The agent is trained for 3e6 steps using the algorithm 1.

The experiments were conducted on a machine equipped with an NVIDIA Quadro RTX 5000. The models were implemented using the PyTorch deep learning framework through the stable baselines libraries, and all experiments were performed on a system running Windows 11. We trained the different instance-size group agents on a fixed number of $3 \times 10^6$ steps. In Fig. 4, we report the training and deployment times of the agents for varying instance sizes.

In Fig. 5, we examine the training dynamics of the agent. To illustrate the agent's learning behaviour, we focus on a middle-sized instance, "ta41" (30 × 20), as a representative middle ground. Six metrics are selected to analyse the training process: episode mean length, mean reward, approximate KL divergence, entropy loss, value function loss and overall loss. The episode mean length and reward indicate overall training performance, while the KL divergence is employed to evaluate training stability and the entropy to analyse the exploration–exploitation tendency.

The episode length is one of the most critical metrics in this research, as it directly translates to the number of agent–environment interactions needed to solve the JJSP. In sub-figure (a), we observe a consistent, monotonic decrease in the episode length. The absence of plateaus indicates the agent is not stuck in suboptimal policies, affirming the effectiveness of the reward function in guiding the agent. During the later stages of training, the episode length stabilizes, suggesting that the agent converges on a consistent policy. The episode reward also exhibits a monotonic rise, confirming the tendency of the agent to learn a better policy, resulting in a better reward collection.

The approximate KL (Kullback–Leibler) divergence measures the difference between the new and old policies. It helps assess one key property of the proximal policy optimization algorithm: ensure that the policy update is not too drastic, preventing instability. In the sub-fig (c) KL-loss curve, we notice that the value spikes at the beginning indicate an aggressive policy update, which correlates with the exploration phase. Then, the KL-loss value decreases, confirming the relatively small and stable policy updates. The entropy loss represents the randomness in the policy. While high entropy implies exploration, low entropy suggests exploitation. The steady rise of the entropy loss in sub-fig (d) during the agent training indicates that the policy is becoming more and more deterministic as the agent gradually shifts to exploitation.

Finally, the value function loss and overall loss indicate the training performance. In an actor–critic setup, the value function reflects the critic's network ability to predict the state value required to calculate the action advantage. This advantage is then used to update the actor network's parameters. Both sub-figures (e) VF-loss and sub-figures (f) PPO-loss show a monotonically decreasing value function loss, confirming that the agent critic is correctly predicting the state value. This positively affects the actor's performance in taking actions that maximize rewards.

In conclusion, key observations emerge from the training performance graphs. The episode's mean length consistently decreases, signifying effective learning. The absence of plateaus indicates the agent avoids suboptimal policies, and the episode reward shows a continuous, monotonic rise, reflecting ongoing improvement in policy learning. Approximate KL divergence initially spikes during exploration but stabilizes later, suggesting consistent, small policy updates. The entropy loss steadily rises, indicating a shift towards a deterministic policy. The monotonic decrease of actors and the critic networks' training losses further confirms the overall training performance.

### 6.2. Experimental results and analysis

Even though finite optimization offers exact solutions, their exorbitant computing cost renders them infeasible to solve bigger problems, paving the way for alternative, less optimal solutions ranging from domain-specific heuristics to higher-level strategy metaheuristics and iterative approaches. In this section, we compare our results with a wide range of approaches, ranging from heuristics and metaheuristics to learning-based algorithms. Using the Taillard instance across all the competing algorithms provided a solid benchmark baseline. The list of competing algorithms can be found in Table 1.

We assess our findings by comparing results with four metaheuristic benchmarks: TMIIG (Tabu-mechanism with the Improved Iterated Greedy algorithm), an enhanced variant within the tabu search family; CQGA (Coevolutionary Quantum Genetic Algorithm), a fusion of evolutionary and genetic algorithms; HGSA, a blend of genetic algorithms with the simulated annealing heuristic; and GA–TS, a hybrid genetic algorithm and Tabu search aiming to leverage the strengths of both approaches. Then, our examination extends to more closely related iterative learning-based methodologies. We juxtapose our findings with three reinforcement learning frameworks: GIN (Graph Isomorphism Network), employing a graph neural network approach. GAM (Gated-Attention Model) employs an attention mechanism-based approach. Finally, DGERD (Disjunctive Graph Embedded Recurrent Decoding Transformer) adopts a transformer-based approach.

The comparison results between our approach and DRL and metaheuristics optimizer are summarized in Table 2. This comparison was conducted on 16 Taillard instances with increasing complexity, ranging from 15 machines × 15 jobs to 100 machines × 20 jobs. The makespan is used as the performance assessment metric for all algorithms. The algorithm with the best performance, resulting in a lower makespan for each instance, is denoted in bold in the table. The optimality gap is calculated based on the best runner-up as a baseline. The optimality equation is given by:

$$\text{Optimality Gap} = -\frac{(C_{\max} - C_{\max(\text{baseline})})}{C_{\max(\text{baseline})}} \quad (8)$$

The findings presented in Table 2 underscore the consistent competitiveness or superiority of PetriRL over alternative methods across all instances evaluated. When contrasting with solutions derived from DRL, PetriRL showcases an optimality gap ranging from 1.1% in the "ta21" instance to 11.6% in "ta01". Notably, the attention-based solution GAM outperformed PetriRL in the "ta71" instance with a 5.6% gap (see Fig. 6).

In comparison to metaheuristic approaches, PetriRL demonstrates superior performance across various instances, exhibiting gaps ranging from 3.3% to 36.3%. While our solution surpassed most metaheuristic competitors, the combination of tabu search and genetic algorithm (TS-GA) achieved better results on instances "ta01" and "ta71." However, it is worth noting that while TS-GA may offer good performance in specific instances, it typically demands extensive fine-tuning and retraining for new instance sizes, unlike the adaptable nature of PetriRL.

**Table 1**
Contending algorithms.

| | |
|---|---|
| **Heuristics** | |
| SPT | Job with shortest processing time |
| LPT | Job with longest processing time |
| SPS | Job with shortest processing sequence |
| LPS | Job with the longest processing sequence |
| SSO | Job with the shortest time of subsequent operation |
| LSO | Job with the longest time of subsequent operation |
| **Metaheuristics** | |
| TMIIG | Tabu with the improved iterated greedy algorithm |
| CQGA | Coevolutionary quantum genetic algorithm |
| HGSA | Genetic algorithm combined with simulated annealing |
| GA–TS | Hybrid genetic algorithm and tabu search |
| **Learning-based** | |
| GIN | Graph Isomorphism Network |
| GAM | Gated-Attention Model |
| DGERD | Disjunctive Graph Embedded Recurrent Decoding Transformer |

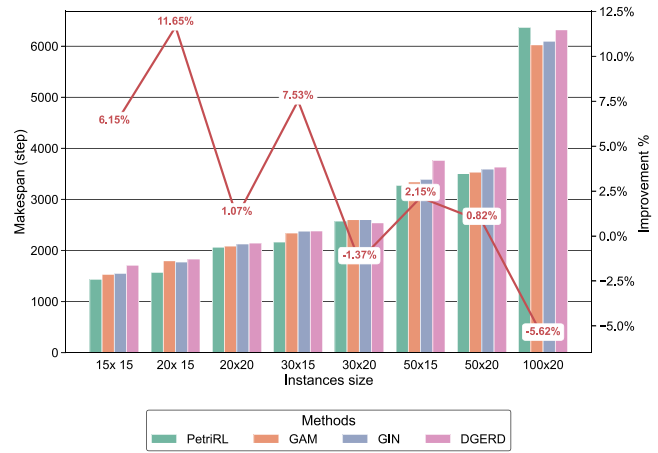

**Fig. 6.** Comparison of the experimental results using PetriRL and learning-based approaches. The barplot is the makespan using different algorithms for different instance sizes. The associated *y*-axis is on the left. In the line plot is the improvement in percentage when using PetriRl compared to the second-best performer. The associated *y*-axis is on the right.

**Table 2**
Comparison of Makespan using PetriRL versus Deep reinforcement learning optimizers and Metaheuristic Algorithms.

| Inst | Size | Deep reinforcement learning | | | | | Metaheuristics | | | | | |
|---|---|---|---|---|---|---|---|---|---|---|---|---|
| | | DGERD | GIN | GAM | PetriRL | Gap | TMIIG | CQGA | HGSA | TS-GA | PetriRL | Gap2 |
| ta01 | 15 × 15 | 1711.07 | 1547.4 | 1530.1 | **1436** | 6.1% | 1486 | 1486 | 1324 | **1282** | 1436 | −12.0% |
| ta11 | 20 × 15 | 1833 | 1774.7 | 1797.9 | **1568** | 11.6% | 2011 | 2044 | 1713 | 1622 | **1568** | 3.3% |
| ta21 | 20 × 20 | 2145.63 | 2128.1 | 2086.3 | **2064** | 1.1% | 2973 | 2973 | 2331 | 2331 | **2064** | 11.5% |
| ta31 | 30 × 15 | 2382.63 | 2378.8 | 2342.3 | **2166** | 7.5% | 3161 | 3161 | 2731 | 2730 | **2166** | 20.7% |
| ta41 | 30 × 20 | **2541.22** | 2603.9 | 2603.5 | 2576 | −1.4% | 4274 | 4274 | 3198 | 3100 | **2576** | 16.9% |
| ta51 | 50 × 15 | 3762.6 | 3393.8 | 3343.8 | **3272** | 2.1% | 6129 | 6129 | 4105 | 4064 | **3272** | 19.5% |
| ta61 | 50 × 20 | 3633.48 | 3593.9 | 3534 | **3505** | 0.8% | 6397 | 6397 | 5536 | 5502 | **3505** | 36.3% |
| ta71 | 100 × 20 | 6321.22 | 6097.6 | **6027.1** | 6366 | −5.6% | 8077 | 8077 | 5964 | **5962** | 6366 | −6.8% |

**Table 3**
Comparison of Makespan using PetriRL and various Heuristics on Taillard instances and randomly generated instances.

| Size | Mks/time | Taillard instances | | | | | | | | Random instances | | | | | | | |
|---|---|---|---|---|---|---|---|---|---|---|---|---|---|---|---|---|---|
| | | SPT | LPT | SPS | LPS | SSO | LSO | PetriRL | Gap | SPT | LPT | SPS | LPS | SSO | LSO | PetriRL | Gap |
| (15, 15) | Steps | 1543 | 1909 | 1664 | 1664 | 1598 | 1760 | **1436** | 6.9% | 1709 | 1644 | 1718 | 1718 | 1547 | 1621 | **1337** | 13.6% |
| | Seconds | 0.23 | 0.22 | 0.21 | 0.21 | 0.21 | 0.23 | 0.81 | | 0.17 | 0.16 | 0.17 | 0.16 | 0.18 | 0.25 | 0.58 | |
| (20, 15) | Steps | 1984 | 1953 | 2160 | 2160 | 1941 | 2218 | **1568** | 19.2% | 2090 | 2098 | 2287 | 2287 | **1865** | 2176 | 1870 | −0.3% |
| | Seconds | 0.27 | 0.3 | 0.32 | 0.32 | 0.36 | 0.38 | 1.12 | | 0.4 | 0.55 | 0.42 | 0.37 | 0.31 | 0.31 | 0.84 | |
| (20, 20) | Steps | 2533 | 2447 | 2581 | 2581 | 2233 | 2281 | **2064** | 7.6% | 2436 | 2166 | 2181 | 2181 | 2213 | 2147 | **2098** | 2.3% |
| | Seconds | 0.48 | 0.5 | 0.52 | 0.63 | 0.52 | 0.59 | 1.51 | | 0.68 | 0.54 | 0.4 | 0.41 | 0.58 | 0.54 | 1.12 | |
| (30, 15) | Steps | 2904 | 2377 | 2584 | 2584 | 2585 | 2646 | **2166** | 8.9% | 2507 | 2599 | 2621 | 2621 | 2407 | 2671 | **2356** | 2.1% |
| | Seconds | 0.73 | 0.73 | 0.71 | 0.66 | 0.61 | 0.69 | 1.75 | | 0.47 | 0.51 | 0.62 | 0.67 | 0.57 | 0.82 | 1.3 | |
| (30, 20) | Steps | 3099 | 2927 | 3309 | 3309 | 2642 | 3288 | **2576** | 2.5% | 2957 | 2812 | 3193 | 3193 | 2600 | 2733 | **2577** | 0.9% |
| | Seconds | 0.78 | 0.87 | 1.08 | 1.08 | 0.78 | 0.96 | 2.51 | | 1.21 | 1.2 | 0.98 | 0.88 | 0.85 | 0.78 | 1.89 | |
| (50, 15) | Steps | 3824 | 3602 | 3743 | 3743 | 3710 | 3934 | **3272** | 9.2% | 3775 | 3516 | 3718 | 3718 | 3385 | 3768 | **2599** | 23.2% |
| | Seconds | 1.12 | 1.51 | 1.7 | 1.71 | 1.57 | 1.59 | 4.02 | | 2.01 | 1.24 | 1.13 | 1.33 | 1.51 | 2.43 | 2.71 | |
| (50, 20) | Steps | 4587 | 4032 | 3953 | 3953 | 3692 | 3922 | **3505** | 5.1% | 4230 | 3785 | 4277 | 4277 | 4021 | 4320 | **3552** | 6.2% |
| | Seconds | 2.63 | 2.11 | 2.33 | 2.26 | 1.91 | 2.13 | 4.35 | | 2.19 | 1.98 | 2.45 | 1.99 | 2.51 | 2.14 | 4.18 | |
| (100, 20) | Steps | 6820 | 7039 | 6930 | 6930 | 6723 | 7617 | **6366** | 5.3% | 6901 | 6914 | 6757 | 6757 | 6602 | 7207 | **6432** | 2.6% |
| | Seconds | 8.06 | 7.78 | 8.45 | 9.86 | 7.37 | 10.34 | 11.02 | | 8.12 | 6.58 | 7.79 | 8.23 | 7.33 | 9.11 | 10.86 | |

After evaluating our results against established benchmarks in Deep Reinforcement Learning and metaheuristic optimizers outlined in existing literature, we expanded our testing scope by comparing our results to the performance of heuristics on both Taillard instances and randomly generated ones. Our investigation centred on six heuristics: SPT, LPT, SPS, LPS, SSO, and LSO. These heuristic algorithms interact with the PetriRl environment, making decisions instead of the RL agent. Furthermore, we recorded the inference time across different instance sizes. The corresponding results are in Table 3.

To ensure that our findings can be replicated, we followed the instructions outlined in the original manuscript by E. Taillard titled "Benchmarks for basic scheduling problems" to generate random instances with various sizes. We first start by generating a pseudo-random number $0 < U(0, 1) < 1$ using a linear congruential generator (LCG). We then used U(0,1) to obtain the random numbers in the range [a,b]. We utilized two seeds to generate the random processing times and machine ordering: time-seed = 840 612 802 and machine-seed = 398 197 754. The algorithm used to create the random processing times and machine sequence is detailed in algorithm 2.

Consistent with the comparisons made against both DRL and metaheuristic approaches, the findings presented in Table 3 highlight the robust performance of PetriRL across instances of varying sizes, particularly in terms of minimizing makespan compared to heuristics. The observed optimality gaps range from 5.1% to 19.2% on Taillard instances and from 0.9% to 23.2% on randomly generated instances. It is noteworthy that PetriRL generally requires more inference time than heuristics. Heuristic methods leverage simple rule-based approaches, enabling quick solution generation, especially in smaller instance sizes. This efficiency arises because heuristics bypass the complexities associated with propagating observation vectors through multiple layers of policy networks, a process inherent to DRL. The computational demands of DRL increase with the network size, making heuristics

**Algorithm 2** Random instances generator using LCG [45]

1: **Linear Congruential Generator (LCG):**
2: Constants: $a = 16807$, $b = 127773$, $c = 2836$, $m = 2^{31} - 1$
3: Calculate $k = \left\lfloor \frac{X_i}{b} \right\rfloor$
4: Update the seed: $X_{i+1} = a \cdot (X_i \mod b) - k \cdot c$
5: **if** $X_{i+1} < 0$ **then**
6:     Let $X_{i+1} = X_{i+1} + m$
7: **end if**
8: Generate the pseudorandom number: $u(0, 1) = \frac{X_{i+1}}{m}$
9: Generate the random integer in the range $[a, b]$: $U[a, b] = [a + u(0, 1) \cdot (b - a + 1)]$
10: **Return** $U[a, b]$
11: **Generate the processing time using (use time-seed):**
12: Initialize $d_{ij}$ for all $i$ and $j$: $d_{ij} \leftarrow U[1, 99]$
13: **for** $i = 1$ to $n$ **do**
14:     **for** $j = 1$ to $m$ **do**
15:         $M_{ij} \leftarrow j$
16:     **end for**
17: **end for**
18: **Generate the machining sequence (use machine-seed):**
19: **for** $i = 1$ to $n$ **do**
20:     **for** $j = 1$ to $m$ **do**
21:         Swap $M_{ij}$ and $M_{iU[j,m]}$
22:     **end for**
23: **end for**

comparatively faster. However, as instances become more complex, the discrepancy in computational time between PetriRL and heuristics becomes less apparent. For example, in a (15 × 15) instance, PetriRL required 0.81 s compared to 0.21 s for heuristics. In contrast, in a more complex (100 × 20) instance, PetriRL required 11.02 s compared to an average of 8.46 s for heuristics, representing a less noticeable difference.

In summary, our approach delivers competitive results to various optimization solutions, including heuristics, metaheuristics, and learning-based methods. What sets our approach apart is that it produces good performance and provides a more intuitive and interpretable representation of a job shop. Also, it eliminates the need to create a disjunctive graph for each instance input, which is necessary with competing learning-based approaches and can be tedious.

*6.3. Ablation study*

In this section, we conduct an ablation study to understand the contribution of individual components of the PetriRL framework to the results. We highlight three major components whose combination contributes to the framework's performance: action masking, reward shaping, and event-based control. In Fig. 7, two main environments were considered for the ablation study. First, a relatively small environment in the form of the Taillard Benchmark number "01" representing a JSSP problem of 15 jobs and 15 machines; the second is a larger environment in the form of the Taillard Benchmark number "41" representing a JSSP problem of 30 jobs and 20 machines.

The controlling parameter is the number of agent–environment interactions needed to reach a termination state, which also coincides with the actual length of the episode. Since no maximum number of steps limit was introduced, the only possible termination scenario is delivering all pieces and resolving the JSSP. The decision to employ episode length as a control parameter in the ablation study is deliberate, as it stands as a neutral metric unaffected by the choice of the reward function. Given that the reward function is a component of the ablation study, episode length emerges as a prominent performance criterion for analysis. All the agents were trained for a fixed number of steps, namely $3 \times 10^6$. The evolution of the episode length during the training using the reference setup in the small and large environment, respectively shown in the sub-graphs (a) and (b), can be found in Fig. 7. The reference setup uses a combined masked version of the proximal policy optimization, event-based control, and reward shaping.

To evaluate the impact of the shaped reward function, we replaced the shaped reward with a fixed penalty of minus one for each step, encouraging the agent to minimize the number of steps taken to solve the JSSP. Examining the results in the small environment (c) compared to the reference case (a), we notice that the agent is slowly learning, but the performance is unstable. A comparison of the two sub-graphs reveals that in the reference case, the episode length is monotonically decreasing and stabilizes 500 $10^3$ steps, whereas using the fixed reward results in an unstable evolution of the episode length despite the slow decreasing tendency. The situation in the large environment is different; comparing the sub-graphs (b) and (d), we notice that the agent can no longer learn by using the fixed reward; in this case, reward shaping is vital. This can be explained by the fact that in a small environment, the agent can reach the terminal state without guidance by randomly sampling actions. However, in a larger environment, this becomes highly unlikely. Furthermore, the monotonic rise of the length of episodes in (d) despite the agent's ability to reach termination states indicates that the fixed reward only brings little information about how good or bad the policy is each time step.

After analysing the impact of reward shaping on performance, we explore the benefits of training the agent in an event-based environment. In sub-figures (e) and (f), we analyse the episode length during agent training for small and large instances. In this scenario, the agent–environment interaction is not event-based, meaning the agent must act regardless of the machine's availability. In a small environment, this significantly increases episodes, which can be attributed to poor credit assignment. Without event-based control, the agent will constantly be forced to choose the standby action since no machine is available. This phenomenon is more pronounced in the large instance, as depicted in sub-figure (f), where the agent requires 2700 steps to solve the JSSP, compared to 625 steps in the reference case.

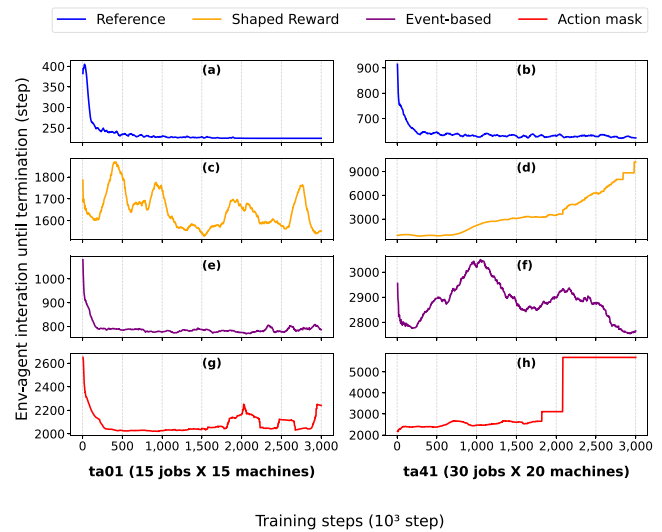

**Fig. 7.** Ablation study conducted using "ta01" (15 × 15) and "ta41" (30 × 20) instances. (a)–(b) is the reference performance using event-based control, action-masking, and reward shaping, (c)–(d) the reward shaping is removed and replaced by a fixed reward, (e)–(f) the event-based control is omitted, (g)–(h) no action making is employed.

After evaluating the impact of event-based control, we examine action masking. Similar to the previous analysis, we investigate the influence of action masking in small and large environments, depicted in sub-graphs (g) and (h), respectively. In the case of a small environment, we observe that the agent can achieve fragile stability. However, it necessitated a remarkably higher 2200 interactions compared to the 225 steps in the reference scenario. This discrepancy can be attributed to the absence of a mask, which increases the likelihood of the agent selecting non-enabled actions, leading to wasted steps. Executing a non-enabled transition maintains the environment in the same state, resulting in a null advantage value. Consequently, the step provides no new information to the agent, hindering policy enhancement. In the larger environment, which results are shown in (h), the side effect of not using a mask is more accentuated. As a consequence, the agent is not able to learn. This is due to the much larger action space and the smaller probability of choosing an enabled action, especially in the early exploration phase.

In summary, the ablation study underscores the significance of action masking, event-based control, and reward shaping within the PetriRL framework. The results show that all three elements are essential pillars for agent training, as removing any one element results in a much higher episode length or renders the agent unable to learn.

*6.4. Generalization*

In this section, we are studying the generalization and flexibility capability of the algorithm. Although the maximum makespan is the main performance metric in the previous paragraph, its dependence on the individual processing times in the different instances is a non-objective metric to compare performances. For example, an instance with long individual operation processing times will most probably end in a longer makespan despite the best agent effort to find the optimal planning. The episode length is chosen as a performance metric to mitigate this problem. Independently of the instance processing time, the episode length can be a good indication of whether the agent policy transfer to another instance is successful or not. The algorithm is put to the test using Taillard Benchmark instances. The instances can be grouped into ten instances with the same characteristics, such as the number of jobs and machines, but different processing time values. To assess the generalization capacity, local and global approaches are tested.

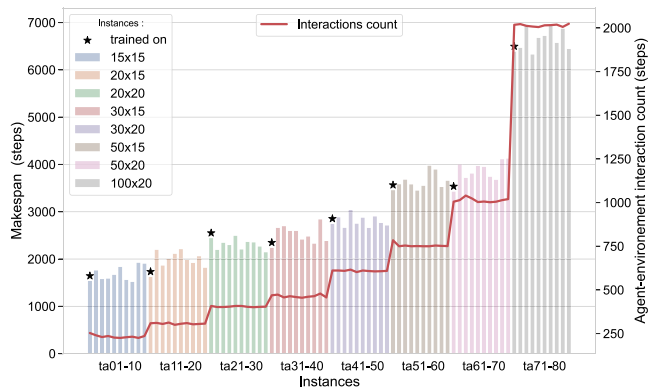

**Fig. 8.** Generability performance on different instances using local-agents. The barplot is the makespan obtained using different agents on different instance sizes, and the line plot is the episode length evolution. Every agent is trained on one instance denoted by a (⋆) and tested on the rest of the instances in the same size group. (For interpretation of the references to colour in this figure legend, the reader is referred to the web version of this article.)

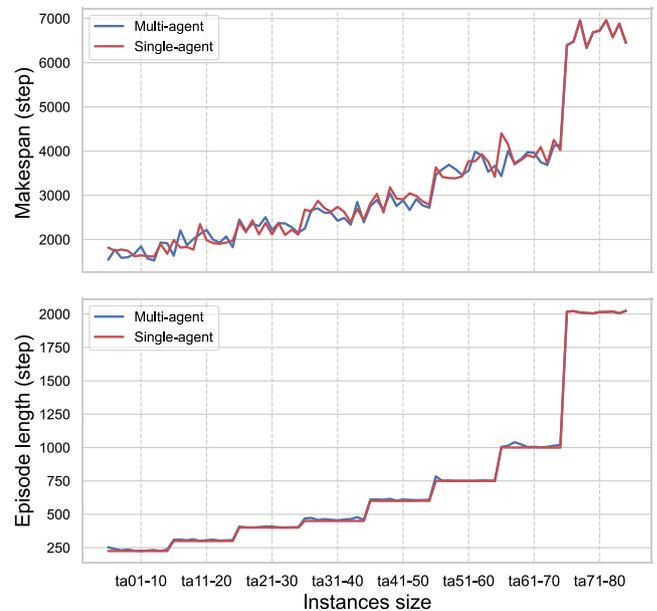

**Fig. 9.** Generability performance on different instances using a global-agent. The agent is only trained on the "ta80" (100 × 20) instance and tested on the rest of the Taillard benchmark instances. The global agent performance in the red line is compared to the local agent performance in the blue line. (For interpretation of the references to colour in this figure legend, the reader is referred to the web version of this article.)

In the local approach, we test generalization capacity on instances of identical sizes where a dedicated agent is trained to solve instances in the same-size group. Shown in Fig. 8 are the results of the generalization capacity for the local agents. The bar plots display the makespan of JSSP instances; instances of identical sizes share the same colour. For example, all instances from "ta01" to "ta10" have 15 jobs × 15 machines. In every size group, the agent was only trained on one of the instances denoted in Fig. 8 by a star on the top. In the second *y*-axis, a line graph (in red) plots the agent–environment interaction count needed to solve the JSSP. We observe the agent's robust, stable performance during deployment across all the instances in every group, shown by the stability in the episode length.

In the global approach, the agent is only trained on one of the largest available instances, in our case, 100 jobs × 20 machines. Once trained, it can handle instances of any size up to 100 jobs × in 20 machines. This capability is possible due to the inherent masking feature of the Petri Net, enabling the use of only a portion of the shop floor. Depending on the instance size, job queues are filled, and the unused job places remain vacant. Consequently, the corresponding selection and allocation transitions are automatically excluded from the action space, thus not affecting the decision-making. Shown in Fig. 9 are the results of the generalization capacity for the global agent, where we contrast the performance of training a dedicated agent for each size group with that of using a single agent across all instances. The comparison reveals a negligible average performance drop of 0.7% compared to using local agents, proving a robust generalization capacity across all the instances.

While the global approach offers undeniable advantages, such as enhanced flexibility, there exist situations where employing local agents proves more advantageous. Training an agent on a larger network comes with a computation price since updating the weights of the larger policy requires more computation. Moreover, to a lesser extent, the performance is also affected during inference since the agent must propagate the input through all the policy network's layers during the forward pass despite most of the action space being masked in the output. In contrast, training a local agent for every size group will result in less computation and deployment time for the smaller instances, as the agent will use a more adapted neural network size. In summary, prioritizing flexibility makes a global agent the better-suited choice. On the other hand, if performance, particularly during inference, is more critical, using a dedicated agent is more efficient.

The proposed framed work is an object-oriented modelling approach, meaning every machine, job, and job is an object. This opens the possibility for the dynamic addition of job operations to in-process jobs or dynamically changing the number of production lines by adding complete jobs requiring new machines, offering great flexibility. During the decision-making, depending on the observation depth, the agent only considers a set number of tokens in the job operation queue to choose an action. This means appending new operations to a job queue is also possible during inference. If the observation depth is one, an operation can be added at every time step without affecting the decision-making. For a single-agent approach, a given number of machines and job places are initially allocated. The system can dynamically adjust the number of utilized resources based on input parameters.

## 7. Conclusion

In this work, we proposed a novel job shop scheduling-solving framework. We employed Petri Net not only as a graphical tool to represent the job shop, promoting explainability, but also as a control tool to govern the autonomous components of the processes. Delegating the non-critical control parts to the Petri net allowed the RL agent to focus on efficient allocation decision-making. The efficiency is further improved by introducing event-based control and action space masking. Thanks to the job shop's main structure in the Petri net model, we could eliminate the need for the laborious preprocessing step of casting the JSSP instances into a disjunctive graph. We compared our performance to a large spectrum of contenders ranging from heuristics and meta-heuristics to learning-based algorithms. The results show that PetriRL consistently achieves competitive or superior solutions compared to other methods. We carried out an ablation study to determine the contribution of individual components of the PetriRL framework to the results, and we found that with varying degrees of importance, all of the action masking, reward shaping, and event-based control are vital for the agent's successful training, especially in larger instances. Finally, we tested the generability and flexibility characteristics of our framework. In the initial phase, individual agents were trained for each size group, resulting in robust generalization within instances of the same size. In the subsequent phase, we extended the framework's capabilities by training a single agent to handle all instances, regardless of size. This leveraged the intrinsic masking capability of Petri Nets through the transition guard function. The across-size generalization demonstrated robust performance with only a minimal drop compared to local generalization, yet it offered greater flexibility. Our framework provides a flexible, efficient, and explainable solution for solving JSSPs.

While strides were made using Petri nets as an interpretable graphical representation, the inherent opacity of the decision-making process within the policy neural network leaves room for improvement.

**CRediT authorship contribution statement**

**Sofiene Lassoued:** Writing – review & editing, Writing – original draft, Visualization, Validation, Software, Resources, Project administration, Methodology, Investigation, Formal analysis, Data curation, Conceptualization. **Andreas Schwung:** Writing – review & editing, Validation, Supervision, Resources, Methodology, Formal analysis, Conceptualization.

**Declaration of competing interest**

The authors declare that they have no known competing financial interests or personal relationships that could have appeared to influence the work reported in this paper.